\documentclass[sigconf]{acmart}
 \usepackage{balance}
 
\AtBeginDocument{%
  }

\copyrightyear{2026}
\acmYear{2026}
\setcopyright{cc}
\setcctype{by}
\acmConference[WWW '26]{Proceedings of the ACM Web Conference 2026}{April 13--17, 2026}{Dubai, United Arab Emirates}
\acmBooktitle{Proceedings of the ACM Web Conference 2026 (WWW '26), April 13--17, 2026, Dubai, United Arab Emirates}
\acmPrice{}
\acmDOI{10.1145/3774904.3793055}
\acmISBN{979-8-4007-2307-0/2026/04}

\begin{document}

\title{Trust in One Round: Confidence Estimation for Large Language Models via Structural Signals}


\author{Pengyue Yang}
\affiliation{%
  \institution{The University of Sydney}
  \city{Sydney}
  \country{Australia}}
\email{pyan8493@uni.sydney.edu.au}

\author{Jiawen Wen}
\affiliation{%
  \institution{The University of Sydney}
  \city{Sydney}
  \country{Australia}
}
\email{jwen8784@uni.sydney.edu.au}

\author{Haolin Jin}
\affiliation{%
 \institution{The University of Sydney}
  \city{Sydney}
  \country{Australia}
}
\email{hjin3177@uni.sydney.edu.au}

\author{Linghan Huang}
\affiliation{%
  \institution{The University of Sydney}
  \city{Sydney}
  \country{Australia}
}
\email{lhua5130@uni.sydney.edu.au}

\author{Huaming Chen}
\affiliation{%
  \institution{The University of Sydney}
  \city{Sydney}
  \country{Australia}
}
\email{huaming.chen@sydney.edu.au}
\authornote{Corresponding author}

\author{Ling Chen}
\affiliation{%
  \institution{University of Technology Sydney}
  \city{Sydney}
  \country{Australia}
}
\email{ling.chen@uts.edu.au}

\renewcommand{\shortauthors}{Pengyue Yang et al.}

\begin{abstract}
Large language models (LLMs) are increasingly deployed in domains where errors carry high social, scientific, or safety costs. Yet standard confidence estimators, such as token likelihood, semantic similarity and multi-sample consistency, remain brittle under distribution shift, domain-specialised text, and compute limits. In this work, we present Structural Confidence, a single-pass, model-agnostic framework that enhances output correctness prediction based on multi-scale structural signals derived from a model's final-layer hidden-state trajectory. By combining spectral, local-variation, and global shape descriptors, our method captures internal stability patterns that are missed by probabilities and sentence embeddings. 
We conduct extensive, cross-domain evaluation across four heterogeneous benchmarks—FEVER (fact verification), SciFact (scientific claims), WikiBio-hallucination (biographical consistency), and TruthfulQA (truthfulness-oriented QA). Our Structural Confidence framework demonstrates strong performance compared with established baselines in terms of AUROC and AUPR. More importantly, unlike sampling-based consistency methods which require multiple stochastic generations and an auxiliary model, our approach uses a single deterministic forward pass, offering a practical basis for efficient, robust post-hoc confidence estimation in socially impactful, resource-constrained LLM applications.

\end{abstract}





\maketitle

\section{Introduction}

Large language models (LLMs) now serve as core components in search engines, dialogue agents, scientific assistants, and decision-support systems, and are increasingly embedded into developer-facing workflows and agentic toolchains (e.g., software engineering) \cite{jin2024llms}. However, despite their impressive generative fluency, LLMs frequently produce incorrect or fabricated content, commonly referred to as hallucinations~\cite{ji2023survey,rawte2023survey}. This gap between generative fluency and factual fidelity has made \emph{post-hoc confidence estimation} a critical requirement for safety-sensitive applications, such as fact verification, reasoning, and content moderation.

Existing confidence estimation methods fall into three main families, each exhibiting inherent limitations. (1) Probability-based estimators, such as mean log-probability, token entropy, and calibration-based scores, are known to be unreliable under distribution shift and miscalibration~\cite{guo2017calibration,ovadia2019can}. (2) Semantic-similarity estimators rely on embedding geometry (e.g., Sentence-BERT~\cite{reimers2019sentence}, DeBERTa-based classifiers), whose performance can degrade severely when input text diverges from the model’s pretraining distribution~\cite{bommasani2021opportunities,liang2022holistic}. (3) Multi-sample consistency methods such as SelfCheckGPT~\cite{manakul2023selfcheckgpt} require repeated stochastic sampling and auxiliary natural language inference (NLI) models, resulting in prohibitive latency and computational cost, making them unsuitable for high-throughput or real-time applications. Furthermore, a fourth emerging line of work explores hidden-state geometry signals, which examine the trajectory of token-level activations rather than output probabilities or embedding-space similarity \cite{fort2019deep, zhou2022understanding}.

These three paradigms are not ad hoc, rather, each is designed to optimize a different axis of the confidence–cost trade-off. Probability-based scores are cost-effective because they read off uncertainty from the model's own output distribution, requiring no additional components and incurring only a small constant-factor overhead over normal decoding~\cite{guo2017calibration,ovadia2019can}. Semantic methods leverage pretrained encoders or sentence representations to inject high-level similarity structure, yielding strong in-domain discrimination when the test distribution closely matches the pretraining corpus~\cite{reimers2019sentence,liang2022holistic}. Multi-sample consistency methods such as SelfCheckGPT~\cite{manakul2023selfcheckgpt} deliberately trade latency and compute for robustness: by aggregating agreement patterns across many stochastic generations and auxiliary NLI or QA models, they approximate a reference-augmented verification signal without requiring labeled data. 

However, these design choices create a structural tension for Web-scale deployment. Probability-based estimators inherit the model's miscalibration and become brittle under distribution shift; semantic estimators depend on the stability of an embedding manifold that can collapse on domain-specialized or long-form text; and sampling-based consistency methods such as SelfCheckGPT are inherently expensive because they combine multiple generations with auxiliary models. And existing efforts on hidden-state geometric signals remain sparse and fragmented, and do not provide a unified framework or a robust characterization that generalizes across domains or model families. Consequently, achieving high-throughput deployment requires a confidence signal that (i) respects strict single-pass and model-agnostic constraints, while (ii) remaining robust under domain shift---motivating a shift from probability and semantic spaces to the structural stability of the hidden-state trajectory.

A key limitation shared by these approaches is their reliance on \emph{surface-level signals}, such as token likelihoods, embedding similarity, or cross-sample agreement, while ignoring the internal computational process that produces model predictions. During decoding, an LLM generates a sequence of hidden states that reflect its evolving reasoning dynamics. Recent studies show that these hidden-state trajectories exhibit distinctive structural patterns associated with uncertainty, including activation variance, unstable attention transitions\cite{michel2019sixteen}, and geometric or topological irregularities in the latent space~\cite{valeriani2023geometry, fitz2024hidden}. However, despite these observations, trajectory-level structure remains largely underexplored as a primary confidence signal in its own right.

This paper introduces \textbf{Structural Confidence}, a framework that models \emph{hidden-state structural stability} as an indicator of confidence that is designed to be robust across distributions within the factual and hallucination-style settings we study. Our key hypothesis is that confident predictions correspond to smooth, low-frequency trajectories with compact global geometry, whereas uncertainty manifests as spectral irregularity, sharp local fluctuations, and dispersed or fragmented global structure. By capturing these multi-scale dynamical perturbations, Structural Confidence aims to surface a class of confidence cues that often remain informative when probability-based and semantic-based signals degrade, particularly under domain shift. Our contributions are as follows:

\textbf{A new confidence modality based on hidden-state structural stability.}
We introduce the Structural Confidence framework, a new confidence modality based on hidden-state structural stability. We define trajectory-level structural descriptors, covering spectral stability, local variation, and global shape coherence of token-level hidden states, as a theoretically motivated and empirically validated confidence signal that robustly complements existing likelihood-, embedding-, and sampling-based estimators.

\textbf{A model-agnostic, single-pass confidence estimator.}
The core estimator operates on structural features extracted from a \emph{single} deterministic forward pass, requiring no sampling, auxiliary models, or architecture-specific internals. We further show that fusing these structural signals with a sentence-level semantic representation yields a lightweight, generalizable estimator, without requiring any extra LLM calls.

\textbf{Extensive and cross-domain evaluation.}
We conduct extensive, cross-domain evaluation across FEVER~\cite{thorne2018fever}, SciFact~\cite{wadden2020fact}, WikiBio~\cite{lebret2016neural}, and TruthfulQA~\cite{lin2022truthfulqa}. Structural Confidence achieves state-of-the-art performance in terms of AUROC and AUPR, while demonstrating robustness under domain shift. Specifically, across the mixed-domain benchmark, our approach is significantly more efficient, for example, the NLI-based SelfCheckGPT consistency baseline incurs on average 5--6$\times$ higher FLOPs and 4--5$\times$ higher latency than our method in the confidence-estimation stage.

Together, these findings suggest that hidden-state structural stability is a promising and underexplored basis for post-hoc confidence estimation within the factual and hallucination-style benchmarks we study. Structural Confidence offers an efficient, model-agnostic, and practically domain-robust solution for assessing LLM reliability, and points toward a line of work in trustworthy AI centred on the internal dynamics of model activations.

\section{Related Work}
Post-hoc confidence estimation for LLMs has been approached through three main paradigms: probability-based estimators, semantic-similarity estimators, and sampling-based consistency methods. A separate line of research examines the geometry and topology of hidden states, providing the theoretical substrate for this work.

\textit{Probability-based confidence estimation.}
Early approaches measure uncertainty via token-level statistics such as mean log-probability, entropy, or temperature-scaled confidence \cite{guo2017calibration, ovadia2019can}. Despite their simplicity, these signals are known to be poorly calibrated, highly sensitive to distribution shift, and often misaligned with correctness \cite{kadavath2022language}. Classical epistemic–aleatoric separation approaches such as Prior Networks require explicit access to model logits and are thus incompatible with our black-box setting~\cite{malinin2018predictive}. Recent studies further show that log-probability can systematically assign high scores to hallucinated or stylistically fluent text \cite{ji2023survey}. Our results confirm these limitations: probability-based baselines collapse under SciFact's domain drift.

\textit{Semantic embedding–based confidence estimation.}
A second paradigm relies on semantic similarity or embedding geometry to assess confidence, including selective prediction~\cite{kamath2020selective}, representation-based analysis of embedding stability~\cite{kovaleva2019revealing,mosbach2020stability}, and sentence-level embedding classifiers~\cite{reimers2019sentence}. These methods perform well in-domain but degrade sharply when the test distribution diverges from the pretraining corpus~\cite{bommasani2021opportunities, liang2022holistic}.

\textit{Sampling-based consistency methods.}
Multi-sample agreement aims to estimate confidence by testing whether a model reaches consistent conclusions across repeated stochastic generations. Reasoning-based hallucinations have been studied extensively, particularly in chain-of-thought settings~\cite{lyu2023faithful}.
The most prominent example is SelfCheckGPT~\cite{manakul2023selfcheckgpt}, which combines multiple GPT samples with auxiliary NLI- or question-generation--based checks. In its original formulation, SelfCheckGPT requires repeated decoding and additional classifier passes, which can be costly and brittle for short or noisy outputs.
In our experiments, we use the official SelfCheckNLI checker from the SelfCheckGPT library as a single-pass baseline: we score a single deterministic generation using a DeBERTa-based NLI model (no multi-sample decoding). This setting isolates the cost and behaviour of the NLI-based checking component, enabling a cleaner comparison to other single-pass estimators. 

\textit{Hidden-state geometry and structural signals.}
A complementary line of work investigates the internal geometry and dynamics of transformer hidden states. Layer-wise probing has shown that intermediate activations encode rich linguistic and structural information~\cite{tenney2019bert}, while mechanistic studies reveal that specific subspaces and feed-forward ``key–value memories'' implement interpretable computation patterns~\cite{geva2021transformer}. More recently, several papers have linked uncertainty to characteristic behaviours in activation space, including increased variance, curvature changes, and unstable attention dynamics~\cite{fort2019deep, zhou2022understanding}. Topological analyses further suggest that latent representations exhibit persistent homological structure that can fragment under distributional stress~\cite{fitz2024hidden}. A related line of work examines the geometry of encoder-only transformers such as BERT and RoBERTa, showing that their hidden states form structured subspaces and evolve along smooth, interpretable trajectories~\cite{valeriani2023geometry, ethayarajh2019contextual}. However, these studies remain primarily diagnostic: they characterise hidden-state phenomena but do not convert trajectory-level structure into a practical, end-to-end confidence estimator under strict single-pass constraints.

\textit{Positioning of this work.}
Our approach builds upon a foundation with prior hidden-state and topology-based analyses, but diverges significantly in objective. Instead of using geometry or topology only for post-hoc interpretation, we \emph{operationalise} trajectory structure as a confidence signal: Structural Confidence extracts a compact set of spectral stability, local variation, and shape-coherence descriptors from the final-layer trajectory, which are then fed into a lightweight supervised estimator. It offers advantages over existing baselines. Our method is deployed as a model-agnostic, drop-in scoring layer that (i) relies solely on features available from a single deterministic forward pass, (ii) assumes no access to gradients, attention maps, or auxiliary models. This work represents the first systematic effort to operationalize hidden-state structural stability as a unified, domain-general modality for post-hoc confidence and rigorously evaluate it against current state-of-the-art methods across heterogeneous factuality tasks.

\section{Problem Definition}
\label{sec:problem}

Confidence estimation for LLMs aims to determine whether a generated answer is likely to be correct. Formally, given an input $x$ and the LLM's single-pass autoregressive output $y = (y_1,\dots,y_T)$ with final-layer hidden states $h_1,\dots,h_T$, the objective is to estimate
\[
    c(x,y) \approx \Pr(\text{correct}(y)\,|\,x,y,h_{1:T}).
\]
In this work, we focus on the deployment regime characterized by strict computational constraints, where the confidence estimator must (i) rely on only a \emph{single deterministic forward pass} of the LLM, (ii) use no auxiliary task-specific models such as NLI rerankers, and (iii) perform no multi-sample decoding such as SelfCheckGPT-style sampling. These constraints reflect the latency, cost, and scalability requirements of Web-scale LLM deployment. Existing confidence estimators fall into three dominant paradigms, yet each conflicts with this budget. Probability-based heuristics (token log-probabilities, predictive entropy) adhere to the single-pass constraint but are known to be miscalibrated and brittle under distribution shift~\cite{guo2017calibration}. Semantic embedding methods estimate confidence from representation similarity, but their geometry often collapses for out-of-domain (OOD) inputs where embeddings drift from their pretraining manifold~\cite{reimers2019sentence}. Sampling-based consistency methods improve robustness by aggregating multiple samples~\cite{manakul2023selfcheckgpt}, but they incur a multi-fold increase in compute and latency and therefore explicitly fall outside the single-pass deployment design.

We posit that these collective limitations arise because existing approaches operate either in the \emph{probability space} or the \emph{semantic space}, both of which are surface-level and sensitive to lexical or distributional mismatch. In contrast, uncertainty fundamentally emerges from the LLM’s \emph{internal computation dynamics}. When the model is confident, its hidden-state trajectory tends to evolve smoothly along a coherent direction; when it is uncertain, the trajectory exhibits local irregularity, higher-frequency oscillation, and global shape inconsistency. These structural signatures are intrinsic to model computation and empirically more stable across domains than probabilities or static embeddings.

We therefore define a third modality for confidence estimation:
\[
    \textbf{structural stability of the hidden-state trajectory}.
\]
Let $\tau = (h_1,\dots,h_T)$ denote the final-layer trajectory. We seek a function
\[
    c_{\mathrm{struct}}(x,y) = g(f(\tau)),
\]
where $f$ extracts multi-scale structural descriptors (local variation, spectral stability, and shape coherence), and $g$ maps them to a scalar confidence score. This formulation reframes confidence estimation as a structural inference problem grounded in the geometry and stability of internal LLM dynamics, and it directly motivates the Structural Confidence framework introduced in Section~\ref{sec:method}.

\section{Method}
\label{sec:method}

Structural Confidence is a single-pass, generator-agnostic confidence estimator designed for online large language model (LLM) services. Its central idea is to extract structural stability signals from a \emph{proxy} hidden-state trajectory computed by a frozen encoder, rather than requiring logits, sampling, or internal activations from the production LLM. This design is explicitly motivated by practical deployment constraints: modern Web APIs (e.g., GPT-4o) expose only text outputs, enforce strict rate limits, and disallow access to token-level probabilities or hidden states. Consequently, any post-hoc confidence estimator must operate under a \emph{single-pass}, \emph{black-box}, and \emph{computationally bounded} regime. Structural Confidence is built around these constraints and aims to provide a stable, efficient, and generalizable mechanism for estimating correctness across diverse Web domains.

\begin{figure*}[t]
    \centering
    \includegraphics[width=0.9\linewidth]{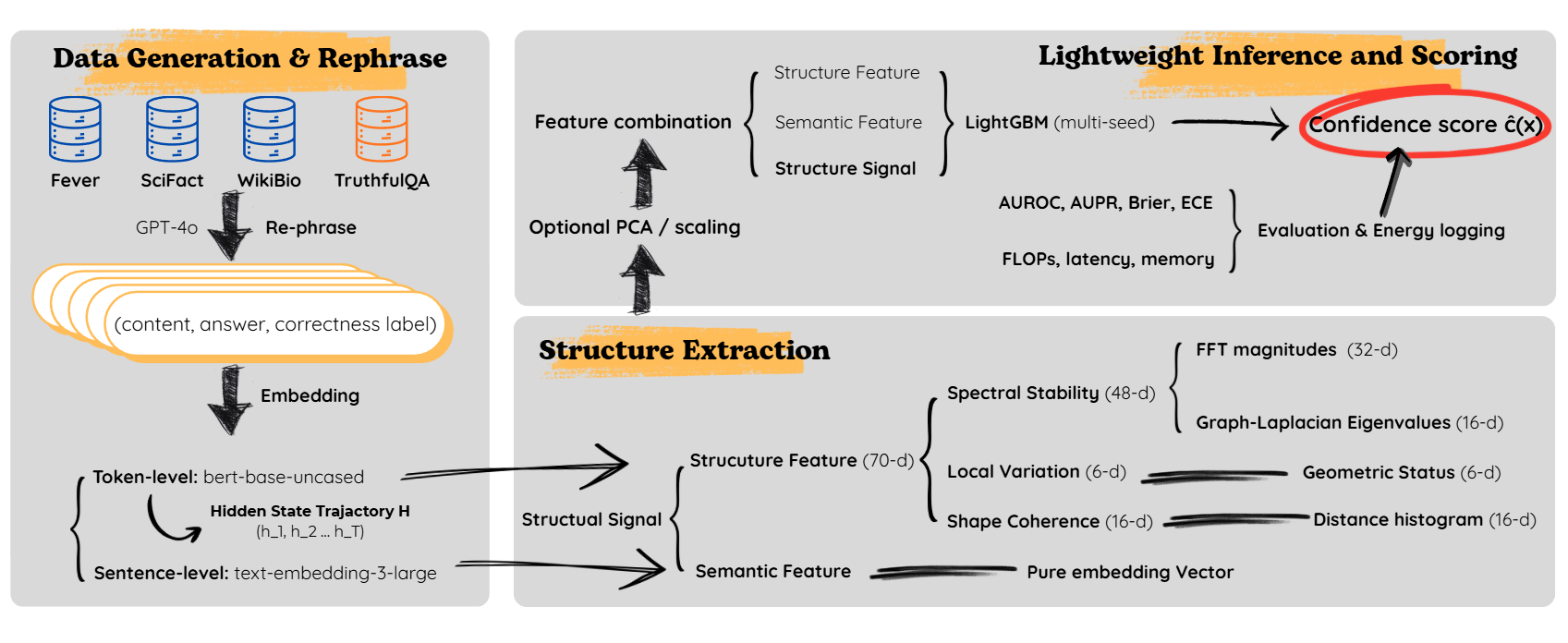}
    \caption{Overall Structural Confidence pipeline. An LLM produces a single deterministic answer; an encoder maps the (context, answer) to a hidden-state trajectory from which multi-scale structural descriptors are extracted and scored by a lightweight confidence model.}
    \label{fig:pipeline}
\end{figure*}

\subsection{Proxy Hidden-State Trajectory Extraction}

Because modern Web LLM APIs do not expose internal hidden states, we construct a \emph{proxy} trajectory using a frozen encoder-only model. Given $(x,\hat{y})$, we concatenate them into $t=\text{concat}(x,\hat{y})$ using fixed delimiters and feed $t$ into a 12-layer transformer encoder with hidden size $768$ instantiated from \texttt{bert-base-uncased}~\cite{devlin2019bert}. The final-layer hidden-state sequence is
\[
H=(h_1,\ldots,h_T) \in \mathbb{R}^{T\times768},
\]
with $T$ truncated or padded to a maximum of $256$ tokens.

\paragraph{Tokenizer robustness.}
Although BERT and GPT-4o use different tokenizers, structural descriptors depend on activation geometry rather than lexical alignment. Empirically, small variations in tokenization do not change the multi-scale patterns encoded in $H$, and the fixed cap of $T=256$ standardizes descriptor computation.

In the complete Structural Signal variant (structure-feature + semantic-feature), we also obtain a sentence embedding $s\in\mathbb{R}^{d_s}$ from a pretrained encoder. No gradients are applied to either encoder, ensuring strict generator-agnostic behaviour.

\subsection{Structural Descriptor Families}

The complete Structural Signal consists of two components: (i) a structure-feature vector derived from the hidden-state trajectory, and (ii) a semantic-feature vector obtained as a fixed sentence embedding $s \in \mathbb{R}^{d_s}$ from a frozen pretrained encoder (OpenAI \texttt{text-embedding-3-large}). The semantic feature serves as a coarse-grained semantic anchor and is concatenated with the structural descriptor.

This section therefore discusses only the structure-feature component. Given a trajectory $H=(h_1,\ldots,h_T)$, we extract three complementary families of structural signals. Each family is length-agnostic: it produces a fixed-size descriptor independent of $T$, enabling consistent inference across Web inputs.

\subsubsection{Spectral stability (48-d).}
This family summarizes how smoothly the hidden-state trajectory evolves. It consists of two components:

\paragraph{Frequency-domain smoothness (32-d).}
We apply a real-valued DFT along the token axis of $H$ and retain the mean and maximum power of the lowest $K=16$ non-trivial frequencies, yielding $2K=32$ features. Zero-padding is applied in token-space before the DFT; because only the low-frequency band is retained, padding does not affect the retained spectral energy.

\paragraph{Graph-spectral diffusion (16-d).}
We construct a similarity graph over tokens and compute the smallest $16$ eigenvalues of the normalized Laplacian $L$. These eigenvalues characterize global smoothness and diffusion along the trajectory. With $T$ capped at $256$, computing the lowest eigenvalues remains tractable and significantly cheaper than multi-sample generation.

\subsubsection{Local variation (6-d).}
Local variation captures short-range instability in the hidden-state trajectory. For each consecutive pair of tokens we compute the displacement $\Delta_t = \lVert h_t - h_{t-1} \rVert_2$ and summarise the trajectory using six statistics: total path length, mean and variance of $\Delta_t$, start–end distance, embedding-wise variance, and centroid norm. These descriptors quantify fine-grained fluctuations that global structural measures may overlook.

\subsubsection{Shape coherence (16-d).}
Global dispersion is summarized by aggregating all pairwise distances $\|h_i-h_j\|_2$ into a 16-bin normalized histogram, producing a length-agnostic "shape signature" that reflects the overall coherence of the trajectory.

\subsection{Unified Structural Representation and Granularity}

Let $f_{\text{spec}}(H)\in\mathbb{R}^{48}$, $f_{\text{loc}}(H)\in\mathbb{R}^{6}$, and $f_{\text{shape}}(H)\in\mathbb{R}^{16}$. Structural Confidence concatenates them into:
\[
u(H)=\big[f_{\text{spec}}(H)\,\Vert\,f_{\text{loc}}(H)\,\Vert\,f_{\text{shape}}(H)\big]\in\mathbb{R}^{70}.
\]

To control structural resolution, we use three granularity modes:

\paragraph{Global.} All features computed on the full trajectory.

\paragraph{Local.} Features computed over overlapping windows ($w=5$, stride $=2$) and averaged across windows. Short windows capture token-level instability that global descriptors may smooth away.

\paragraph{Two-scale.} The global descriptor and the averaged local descriptor are elementwise averaged. Although descriptors differ in nature (spectral, geometric, histogram-based), their normalization ensures comparable scales, and averaging serves as a multi-resolution smoothing analogous to wavelet-based representations.

Unless stated otherwise, the two-scale mode is used as default. Global-only and local-only variants are evaluated in ablations.

\subsection{Confidence Estimator and Training Objective}

We train a gradient-boosted decision tree model (LightGBM) to map either $u(H)$ (Struct-only) or $[u(H)\,\Vert\,s]$ (Struct+Sent) to a correctness probability. A binary logistic objective is used. Trees are shallow and few in number, reflecting the low-dimensional input and the need for fast inference. Hyperparameters (learning rate, number of leaves, feature fraction, max depth) are fixed per dataset and specified in Section~\ref{sec:exp-setup}. Tree-based models provide robustness to heterogeneous feature scales and non-linear interactions between descriptor families.

\subsection{Computational Considerations}

The end-to-end cost per instance is:
\[
\text{Cost}(T) = \text{Cost}_{\text{enc}}(T) + O(T\log T + T^2),
\]
where $\text{Cost}_{\text{enc}}(T)$ is the encoder forward pass. With $T$ capped at $256$, both $O(T\log T)$ FFT features and $O(T^2)$ shape-coherence and graph-spectral components remain CPU-feasible. In practice (Section~\ref{sec:efficiency}), Structural Confidence achieves significantly lower latency and FLOPs compared to sampling-based methods (e.g., 10 rounds of SelfCheckGPT). Moreover, all computations are deterministic and do not require repeated generator calls, making the approach suitable for real-time Web inference.

\subsection{Why a Proxy Encoder is Sufficient}
\label{sec:proxy-validity}

Because proprietary LLMs such as GPT-4o do not expose hidden states, Structural Confidence relies on a lightweight encoder-only transformer to obtain token-level trajectories. Importantly, we do \emph{not} assume that the proxy encoder reproduces the generator’s internal activations. Instead, we adopt an input-conditioned view of activation geometry: many stability- and uncertainty-related patterns are induced by the structure of the context–answer sequence and the autoregressive decoding process, rather than by model-specific parameters.

Evidence for this perspective comes from prior work showing that geometric and structural signatures—such as low-frequency spectral concentration, reduced local curvature, and coherent manifold organization—appear across different transformer families~\cite{morris2023text,hewitt2019structural}. These patterns are therefore at least partly model-agnostic.

Our empirical results further support the adequacy of a proxy encoder. The same structural descriptors computed from BERT trajectories consistently separate correct from incorrect outputs across FEVER, SciFact, and WikiBio-LLM, and transfer smoothly to the out-of-domain TruthfulQA benchmark. Performance degrades gradually rather than catastrophically, suggesting that the proxy trajectory preserves the text-induced stability signals that correlate with correctness.

Thus, while the proxy encoder does not approximate GPT-4o layerwise, it provides a stable and computationally lightweight surrogate trajectory whose geometric and spectral structure retains the correctness-sensitive patterns needed for post-hoc confidence estimation. A direct comparison with true generator trajectories is left for future work.

\subsection{Theoretical Motivation: Why Structural Signals Predict Confidence}
\label{sec:why-structure}

We posit that confidence is reflected in the stability of autoregressive hidden-state dynamics: when the model is confident, the trajectory is smooth and low-variance; under uncertainty or hallucination, it becomes irregular, high-curvature, and structurally fragmented. This behaviour is consistent with how transformer decoders update their hidden states during autoregressive prediction.

At each token step, the model updates its representation via $h_{t} = f(h_{t-1}, \mathrm{Attn}_t, \mathrm{logits}_t)$. Confident predictions typically exhibit peaked next-token distributions and stable attention patterns, so the update varies smoothly across $t$, yielding trajectories with (i) low-frequency spectral energy, (ii) small and homogeneous incremental changes, and (iii) coherent global structure.

Uncertain or hallucinated predictions break this stability. When the model faces ambiguous evidence or conflicting contextual cues, the next-token distribution tends to become less peaked (higher-entropy) and more variable across steps, which amplifies the sensitivity of the attention weights and the MLP pathways. The resulting update function becomes locally chaotic: small variations in logits or attention cause disproportionately large changes in $h_t$. This induces (i) high-frequency spectral components, (ii) irregular local curvature and step-length variance, and (iii) fragmentation and topological noise in the global trajectory.

These effects have been observed empirically across transformer families. Spectral irregularity correlates with uncertainty-driven activation drift~\cite{rahaman2019spectral}, local geometric instability aligns with inconsistent token-level decision boundaries~\cite{mccoy2020berts}, and loss of topological coherence has been linked to unreliable internal reasoning pathways~\cite{fitz2024hidden}. Importantly, these phenomena arise from the dynamics of autoregressive decoding rather than model-specific parameters, making them stable across proxy encoders and domains.

Accordingly, Structural Confidence operationalizes three measurable families—spectral stability, local variation, and global shape coherence—as model-agnostic indicators of trajectory stability: stable trajectories imply high confidence, while unstable trajectories signal representational uncertainty and likely errors.

\section{Experimental Setup}
\label{sec:exp-setup}

We design the experiments to answer three questions: (Q1) How competitive are structural confidence signals under a unified single-pass protocol? (Q2) How robust are these signals under domain shift and mixed-domain training? (Q3) Which structural design choices---signal families, trajectory granularity, or semantic augmentation---are necessary for strong confidence estimation? This section describes the datasets, LLM generation protocol, proxy encoder extraction, confidence estimators, and evaluation metrics. Implementation follows our released codebase exactly, and all results use cached feature files generated by the scripts in Sections~\ref{sec:method} and~\ref{sec:ablation-results}.

\subsection{Evaluation Tasks}
We evaluate confidence estimation in two settings. \textbf{(1) Sentence-level prediction}: given an instance $(x, a)$ consisting of an input claim/question $x$ and a GPT-4o-generated answer $a$, a confidence estimator outputs a scalar score predicting whether $a$ is factually correct. \textbf{(2) Passage-level ranking}: for a set of generations from the same benchmark, we assess whether confidence scores rank correct outputs above incorrect ones. This ranking view is identical to the AUC-style factuality assessment used in recent hallucination and truthfulness work.

\subsection{Datasets}
We follow the LLM-ese construction in Section~\ref{sec:method}. FEVER~\cite{thorne2018fever} and SciFact~\cite{wadden2020fact} contain claims paired with binary support/refute labels. WikiBio~\cite{lebret2016neural} provides infobox-conditioned biography generations annotated as factually correct or incorrect. All three benchmarks are converted into single-pass evaluation sets by prompting GPT-4o once per instance and aligning each instance with a binary correctness label. After removing malformed outputs, the evaluation sets contain 1{,}893 (FEVER), 809 (SciFact), and 780 (WikiBio) instances.

For \textbf{cross-domain evaluation}, we merge the LLM-ese train and dev splits into a mixed-domain training pool (\texttt{mix\_train}, 5{,}455 instances; \texttt{mix\_dev}, 778 instances). The \texttt{mix\_test} split (3{,}131 instances) contains held-out portions of all three datasets plus the entire TruthfulQA benchmark~\cite{lin2022truthfulqa}, excluded from training.

\subsection{LLM Generation Protocol}
Following SelfCheckGPT’s single-pass design, GPT-4o is queried once per instance using a unified instruction template. All outputs are generated deterministically using greedy decoding (temperature $=0$, top-$k=1$). No sampling, resampling, or paraphrasing is used. All confidence estimators operate strictly on these fixed outputs.

\subsection{Proxy Encoder and Structural Feature Extraction}
Because GPT-4o does not expose hidden states, we adopt a lightweight proxy encoder as described in Section~\ref{sec:method}. The concatenated context–answer sequence is encoded with \texttt{bert-base-uncased}; final-layer hidden states $(h_1,\ldots,h_T)$ are extracted from a single forward pass with a maximum length of 256 subword tokens. Structural descriptors are computed exactly as implemented in our extraction scripts: (i) \textit{local variation} (geometric stability), (ii) \textit{spectral stability} (FFT and graph-Laplacian spectra), and (iii) \textit{shape coherence} (distance-histogram and TDA summaries). The concatenation yields a fixed 70-dimensional vector per instance. All datasets use the same configuration.

\subsection{Baselines}
We compare Structural Confidence to four families of post-hoc confidence scores.
\textbf{(1) Probability-based heuristics}: mean log-probability (Mean logp; \cite{guerreiro2023looking}) and predictive entropy computed from token-level log-probabilities using an open-weight model (\texttt{logp\_family}), because GPT-4o does not expose token-level log-probabilities or logits in our black-box API setting. We compute these scores on the same deterministic GPT-4o answer by applying teacher forcing with the open-weight model and extracting per-token log-probabilities.
\textbf{(2) Semantic baseline}: a \textit{Semantic-feature} variant using mean-pooled BERT representations and the same LightGBM classifier, together with a Semantic Entropy (NLI, single-pass) score computed from entailment probabilities on the same deterministic GPT-4o answer; unlike the original formulation~\cite{farquhar2024detecting}, this instantiation operates under a strictly single-pass, black-box setting without additional generations or auxiliary QA models.
\textbf{(3) Multi-sample consistency}: the NLI-based SelfCheckGPT baseline~\cite{manakul2023selfcheckgpt} using the official 'SelfCheckNLI' model. We average contradiction probabilities across passages and invert them to obtain correctness scores. We do not include heavier multi-sample reasoning methods (e.g., Stable Explanations \cite{chen2024training}) because they require 10–50 generations per query and external encoders.
\textbf{(4) Geometry-based outlier baseline}: a $k$-means outlier detector (\texttt{kmeans\_outlier}) that scores each instance by its distance to the nearest centroid in the representation space. This baseline instantiates a standard clustering-based outlier detection heuristic, for which we cite recent comparative evaluations of clustering-based outlier detection methods~\cite{sanchez2025comparative}. We restrict comparison to such single-pass, black-box baselines and do not include white-box internal-state methods that require access to the generator’s hidden layers.
In addition, we include a RACE-style structural baseline in our efficiency comparison, instantiated as a five-module proxy-trajectory pipeline following the original design~\cite{wang2025joint}; this baseline shares the same proxy encoder but replaces our lightweight descriptors with a deeper, multi-stage structural analysis.

\subsection{Supervised Training Protocol}
Supervised estimators (\textit{Structure-feature}, \textit{Semantic-feature}, \textit{Structural Signal}) use LightGBM with binary logistic loss. Training strictly follows our scripts: 200 trees, learning rate 0.05, unlimited depth, and fixed random seeds for reproducibility. No dataset-specific tuning or validation heuristics are used. Single-domain models are trained on each dataset’s LLM-ese train/dev split; cross-domain models are trained on \texttt{mix\_train} and validated on \texttt{mix\_dev}. Test sets remain strictly unseen.

\subsection{Evaluation Metrics}
We report discrimination (AUROC, AUPR) and calibration (Brier Score, Expected Calibration Error) following standard practice in LLM truthfulness~\cite{liu2022token,guo2017calibration,manakul2023selfcheckgpt}. All metrics operate on the continuous scalar confidence score produced by each method.

\subsection{Efficiency Protocol}
We measure the computational cost of the \emph{confidence estimation stage only} using instrumented FLOPs and wall-clock measurements. The shared GPT-4o forward pass is excluded for all methods. For multi-pass baselines such as SelfCheckGPT, we report both the cost of a single-pass NLI check and the extrapolated cost for $S$ hypothetical samples (scaled linearly with $S$ as in~\cite{manakul2023selfcheckgpt}). Structural Confidence uses only a single BERT-base pass and CPU-side feature extraction.

\textit{Hardware setup.}
All experiments were conducted on a single workstation equipped with an AMD Ryzen 9~5900HX CPU, 32\,GB of RAM, and an NVIDIA GeForce RTX~3080 Laptop GPU with 16\,GB of VRAM. 
Unless otherwise stated, all confidence estimation methods share this hardware configuration.
We measure FLOPs and wall-clock latency for the confidence-estimation stage only, excluding the shared GPT-4o generation cost.

\section{Results and Analysis}
We evaluate Structural Confidence in four parts: main sentence-level prediction results, cross-domain robustness, ablation studies of key design choices, and efficiency analysis.

\subsection{Sentence-Level Confidence Prediction}
\label{sec:results-main}
Table~\ref{tab:main_results} compares all confidence estimators across FEVER, SciFact, and WikiBio. We focus on three questions: how structural signals compare to probability-based heuristics, how they behave relative to semantic baselines, and how the fused configuration behaves against strong reference methods such as SelfCheckGPT.

\textit{Structural signals vs probability-based heuristics.}
Mean-log-probability and entropy struggle on FEVER and WikiBio, often falling near or below random-guess AUROC. In contrast, \textit{Structure-feature} provides a stable ranking signal across all three datasets without using token probabilities. On FEVER and WikiBio, \textit{Structure-feature} outperforms both probability-based heuristics in AUROC and AUPR, supporting the premise that trajectory-level stability captures correctness information that simple likelihoods cannot. On SciFact, mean-log-probability is competitive, but all probability-based methods remain brittle and offer only moderate discrimination.

\textit{Structural signals versus semantic representations.}
\textit{Semantic-feature} performs well on FEVER and WikiBio (0.682/0.683 AUROC), but collapses on SciFact (0.255 AUROC), consistent with prior observations that sentence embeddings degrade under domain and stylistic shift~\cite{reimers2019sentence}. Structural signals degrade smoothly rather than catastrophically: \textit{Structure-feature} remains clearly discriminative on FEVER and WikiBio and retains non-trivial signal on SciFact (0.460 AUROC), providing a more conservative confidence estimate when semantic encoders misfire. This pattern is desirable in Web settings where domain and style cannot be controlled.

\textit{Fused configuration versus strong reference baselines.}
The fused configuration \texttt{structure signal} combines the strengths of both sources, achieving top or near-top AUROC/AUPR among single-pass baselines on FEVER and WikiBio and avoiding the embedding collapse on SciFact. Because all supervised variants share the same classifier and hyperparameters, improvements reflect complementary information rather than additional capacity. Compared to SelfCheckGPT (NLI), which attains the strongest AUROC on FEVER (0.700) but is substantially more expensive (Section~\ref{sec:efficiency}), \texttt{Struct+Sent} trades a small accuracy gap on FEVER for a much lower computational footprint, while matching or outperforming SelfCheckGPT on WikiBio in AUPR. For applications where repeated sampling or NLI passes are infeasible, Structural Confidence therefore offers a competitive single-pass alternative.

\textit{Discrimination and calibration.}
Beyond AUROC and AUPR, calibration metrics (Brier Score and ECE) follow the same trend: structural descriptors and the fused \texttt{Struct+Sent} model are consistently better calibrated than probability-only baselines on FEVER and WikiBio, and remain comparable on SciFact. This suggests that structural stability not only improves ranking but also yields confidence scores that better track empirical correctness frequencies.

\begin{table}[t]
\centering
\small
\caption{AUROC / AUPR on FEVER, SciFact, and WikiBio. Bold entries denote the fused \texttt{Structural Signal} configuration.}
\label{tab:main_results}
\begin{tabular}{p{2.5cm}|cc|cc|cc}
\toprule
\textbf{Method} & \multicolumn{2}{c|}{FEVER} & \multicolumn{2}{c|}{SciFact} & \multicolumn{2}{c}{WikiBio} \\
& AUC & AUPR & AUC & AUPR & AUC & AUPR \\
\midrule
Spectral Stability & 0.555 & 0.268 & 0.475 & \textbf{0.725} & 0.504 & 0.309 \\
Shape Coherence & 0.552 & 0.263 & 0.522 & 0.701 & 0.540 & 0.295 \\
Local Variation & 0.534 & 0.247 & 0.515 & 0.689 & 0.565 & 0.320 \\
\midrule
Structure-feature & 0.573 & 0.274 & 0.460 & 0.670 & 0.543 & 0.354 \\
Semantic-feature & 0.682 & 0.349 & 0.255 & 0.602 & 0.683 & 0.430 \\
\textbf{Structural Signal} & \textbf{0.683} & \textbf{0.348} & 0.335 & 0.640 & \textbf{0.686} & \textbf{0.454} \\
\midrule
Mean logp~\cite{guerreiro2023looking} & 0.438 & 0.208 & 0.505 & 0.677 & 0.485 & 0.295 \\
Semantic entropy (NLI, single-pass)~\cite{farquhar2024detecting} & 0.432 & 0.201 & 0.512 & 0.684 & 0.453 & 0.264 \\
KMeans-outlier~\cite{sanchez2025comparative} & 0.518 & 0.234 & 0.544 & 0.696 & 0.594 & 0.355 \\
SelfCheckGPT~\cite{manakul2023selfcheckgpt} & 0.700 & 0.364 & 0.458 & 0.650 & 0.564 & 0.312 \\
\bottomrule
\end{tabular}
\end{table}

\subsection{Cross-Domain Robustness}
\label{sec:crossdomain-results}
\begin{figure}[t]
\centering
\includegraphics[width=\linewidth]{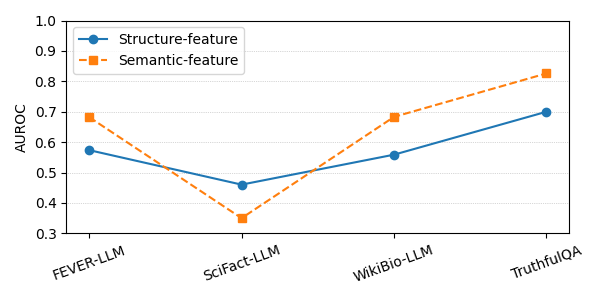}
\caption{Cross-domain AUROC for \textit{Structure-feature} and \textit{Semantic-feature} trained on \texttt{mix\_train} and evaluated on four domains.}
\label{fig:crossdomain_struct_sent}
\end{figure}

We now examine how structural and semantic confidence signals behave under domain shift. A single model is trained on the mixed-domain \texttt{mix\_train} pool and evaluated on FEVER-LLM, SciFact-LLM, WikiBio-LLM, and the out-of-domain TruthfulQA. Figure~\ref{fig:crossdomain_struct_sent} plots AUROC for \textit{Struct-only} and \textit{Semantic-feature} across domains.

\textit{Structural signals generalize smoothly.}
\textit{Structure-feature} demonstrates stable performance across FEVER-LLM, SciFact-LLM, WikiBio-LLM, and TruthfulQA, confirming that structural stability captures domain-invariant regularities in hidden-state dynamics. Despite a marginal degradation in performance when moving from training domains to TruthfulQA, it remains robust, preventing the performance collapse typically seen in cross-domain shifts.

\textit{Semantic signals oscillate strongly.}
\textit{Semantic-feature} varies sharply by domain, with large drops in AUROC on SciFact-LLM and TruthfulQA. This mirrors the single-domain findings and highlights the fragility of sentence embeddings under stylistic and topical shift, especially when training and test domains diverge.

\textit{Complementary fusion.}
When combined, \textit{Structural Signal} tends to track the stronger of structural or semantic signals in each domain, demonstrating that structural stability offers a robust backbone when semantic encoders are misaligned, while semantics act as a strong in-domain refinement when reliable. For Web-scale deployment, where inputs span heterogeneous domains, this combination is strictly safer than relying on embeddings alone.

\subsection{Ablation Studies}
\label{sec:ablation-results}
In this section, we study the main structural design choices for ablation studies, including signal families, semantic augmentation, trajectory granularity, and descriptor dimensionality.

\textit{Structural families are complementary.}
Table~\ref{tab:ablation-families} decomposes the full structural descriptor into its three families. Each family performs best on at least one dataset: spectral stability is strongest on FEVER, shape coherence on SciFact, and local variation on WikiBio. The concatenated \textit{Structure-feature} variant remains close to the best single family on each dataset while being more stable across all of them, indicating that the three families capture complementary aspects of trajectory stability.

\begin{table}[t]
\centering
\small
\caption{Ablation AUROC for structural families and semantic fusion. (\textit{Local}: Local Variation, \textit{Spectral}: Spectral Stability,\textit{Shape}: Shape Coherence, \textit{Struct}: Structure-feature, \textit{Sem}: Semantic-feature)}
\label{tab:ablation-families}
\begin{tabular}{lcccccc}
\toprule
Dataset & Local & Spectral & Shape & Struct & Sem & \textbf{Structural Signal} \\
\midrule
FEVER & 0.538 & 0.579 & 0.553 & 0.574 & 0.682 & 0.683 \\
SciFact & 0.455 & 0.449 & 0.499 & 0.460 & 0.350 & 0.347 \\
WikiBio & 0.570 & 0.532 & 0.520 & 0.559 & 0.683 & 0.685 \\
\bottomrule
\end{tabular}
\end{table}

\textit{Semantic augmentation: when does it help?}
On FEVER and WikiBio, \textit{Semantic-feature} is the strongest single-source signal, and simple concatenation (\textit{Structural Signal}) yields small but consistent gains, suggesting that structural descriptors refine an already strong semantic baseline. On SciFact, however, \textit{Structure-feature} clearly dominates \textit{Semantic-feature}, and naive fusion does not help, showing that semantic information is not uniformly beneficial under domain shift and must be combined carefully.

\textit{Trajectory granularity and dimensionality.}
Varying trajectory granularity (global, local, two-scale) reveals that purely local windows are insufficient on long, noisy claims (FEVER, mixed-domain), where global and two-scale variants provide better AUROC. SciFact, with shorter and more templated claims, benefits more from local structure. PCA compression from 70 to 32/16/8 dimensions yields only minor AUROC changes on FEVER and even small gains on SciFact, indicating that the descriptor is redundant and can be stored in a compact subspace.

\subsection{Efficiency Analysis}
\label{sec:efficiency}
Finally, we compare the compute and latency cost of Structural Confidence against stronger but more expensive baselines. Table~\ref{tab:efficiency} reports relative FLOPs and latency, normalised so that single-pass Structural Signal has cost 1.0×. The importance of minimizing inference-time compute is well documented in recent surveys on LLM deployment under resource constraints~\cite{girija2025optimizing}. All numbers reflect the confidence estimation stage only.

\textit{Single-pass structural confidence is the cheapest.}
Structural Confidence requires a single BERT-base forward pass plus lightweight CPU-side feature extraction and LightGBM inference. This yields the lowest FLOPs and latency among methods that go beyond raw log-probabilities and single-vector sentence embeddings.

\textit{RACE-style and SelfCheckGPT baselines are substantially more expensive.}
RACE-style structural pipelines incur approximately $4\times$ FLOPs and $3\times$ latency relative to the Structural Signal (structure-feature + semantic-feature), while SelfCheckGPT (NLI) incurs roughly $6\times$ FLOPs and $5\times$ latency due to the extra NLI model and longer processing pipeline. These costs would grow linearly with the number of samples in a multi-sample SelfCheckGPT configuration. In contrast, Structural Confidence remains strictly single-pass and proxy-based.

\begin{table}[t]
\centering
\small
\caption{Relative FLOPs and latency of confidence estimators (Structural Confidence = $1.0$).}
\label{tab:efficiency}
\begin{tabular}{lcc}
\toprule
Method & FLOPs & Latency \\
\midrule
Structural Confidence & 1.0 & 1.0 \\
RACE baseline & 4.0 & 3.0 \\
SelfCheckGPT (NLI) & 6.0 & 5.0 \\
\bottomrule
\end{tabular}
\end{table}


\section{Conclusion}

This work introduced \textit{Structural Confidence}, a single-pass, model-agnostic framework for post-hoc confidence estimation in large language models. By analysing the geometric and spectral structure of hidden-state trajectories, and in ablation variants with their topological summaries, it derives stability-based confidence signals that are distinct from traditional approaches, i.e., token likelihoods, sentence-level semantics, and multi-sample consistency scores, requiring only a single encoder forward pass. Using FEVER, SciFact, and WikiBio-hallucination as the benchmarks, Structural Confidence remains reliable under domain shift, outperforms probability-based baselines, and provides complementary gains over sentence-level semantic embeddings. In a mixed-domain regime trained on the union of FEVER-, SciFact-, and WikiBio-LLM and evaluated on these three test sets plus the out-of-domain TruthfulQA benchmark, the same structural model maintains competitive AUROC and AUPR without dataset-specific tuning, and the fused \texttt{Struct+Sent} variant offers a favourable accuracy--efficiency trade-off compared to an NLI-based SelfCheckGPT baseline in the confidence-estimation stage. Our study is limited to a small set of factual and truthfulness benchmarks, one proprietary base model, and FLOPs-/latency-based efficiency estimates on specific hardware configuration. Future work will extend structural confidence modelling to other Web tasks, open-weight and multimodal models, and integration into end-to-end systems such as fact-checking, retrieval-augmented assistance, and content moderation.

\bibliographystyle{ACM-Reference-Format}
\balance
\bibliography{refs}

\appendix

\section{Additional Properties of Structural Descriptors}
\label{app:struct-properties}

This appendix complements Section~\ref{sec:method} by summarising robustness, complexity, and portability properties of the structural descriptors introduced in the main text. We focus on interpretability and reuse, rather than restating algorithmic definitions.

\subsection{Fixed Dimensionality and Length-Agnostic Design}

All descriptor families (local variation, spectral stability, and shape coherence) are designed to have fixed dimensionality, independent of the generation length~$T$. Local variation aggregates token-to-token geometric statistics into a small number of scalars. Spectral descriptors compress the FFT magnitude profile into summaries of low-frequency energy, dominant components, and spectral entropy. Shape coherence builds fixed-bin histograms of pairwise distances, optionally combined with a few topological statistics.

In the global setting used in our experiments, the final structural descriptor therefore has constant dimension, enabling uniform application across Web queries of widely varying length.

\subsection{Robustness to Activation and Configuration Variations}

The descriptors operate directly on raw hidden-state trajectories without explicit per-sample normalisation. Empirically, the use of energy ratios, aggregated geometric statistics, and distance distributions yields robustness to moderate shifts in activation scale and offset.

Across all datasets (FEVER, SciFact, WikiBio, and TruthfulQA), structural features and downstream confidence scores remained numerically stable under multiple random seeds, data splits, and minor configuration variations. This behaviour aligns with stability principles in topological data analysis, where summary functionals (e.g., persistence diagrams) are provably stable under perturbations of the underlying space~\cite{cohen2005stability}. We do not claim new guarantees here, but the same intuition applies: descriptors that focus on global geometric structure are less sensitive to microscopic activation noise.

\subsection{Computational Complexity}

The descriptors are inexpensive relative to the transformer forward pass. Local variation requires only vector differences and simple reductions over~$T$ steps. Spectral descriptors use an $O(T \log T)$ real-valued FFT per trajectory, negligible compared to model inference. Shape coherence uses vectorised pairwise distances and fixed-bin histogramming; in windowed configurations, the computation is restricted to fixed-size chunks, keeping the effective cost close to linear in~$T$ for long sequences.

Consequently, Structural Confidence is dominated by the cost of a single LLM forward pass. The structural computation adds only a small constant factor, quantified empirically in Section~\ref{sec:efficiency} and Appendix~\ref{app:ablation-energy}.

\subsection{Reusability Across Models and Tasks}

Because the descriptors depend solely on the hidden-state trajectory and do not require token identities, logits, or calibration layers, they transfer across decoder-only LLMs without modification. Any model exposing final-layer hidden states can be plugged into the Structural Confidence pipeline, making the descriptors reusable across Web tasks such as fact-checking, scientific assistance, or moderation.

\section{Extended Visualisation of Structural Confidence}
\label{app:visualisation}

To complement the numerical results in Section~\ref{sec:results-main} and Section~\ref{sec:crossdomain-results}, we provide additional visualisations derived from the same result files used in the main tables.

\subsection{Cross-Dataset AUROC Profile}

\begin{figure*}[t]
\centering
\includegraphics[width=0.75\linewidth]{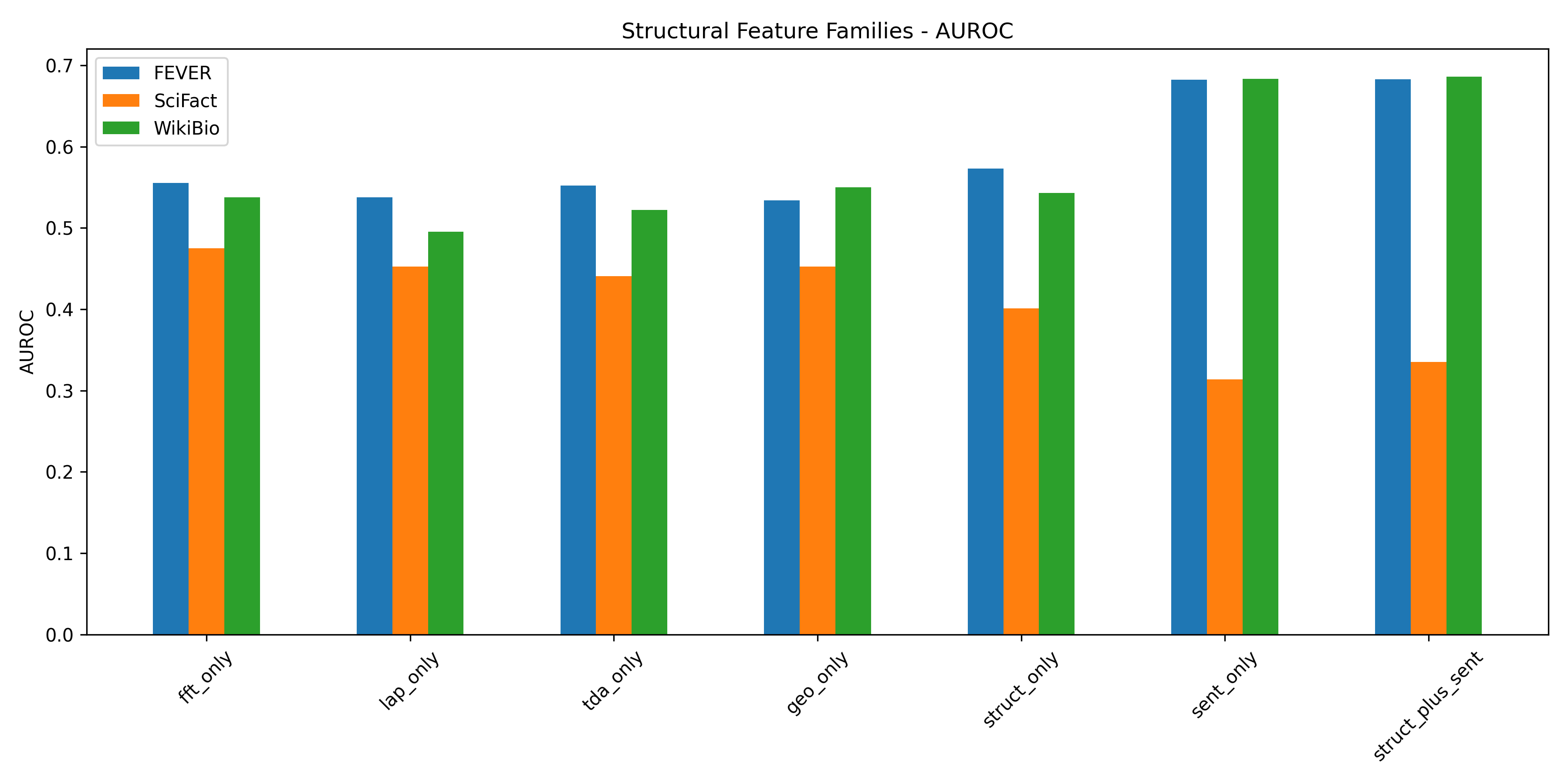}
\caption{AUROC of fft\_only, lap\_only, tda\_only, geo\_only, struct\_only, sent\_only, and struct\_plus\_sent across FEVER, SciFact, and WikiBio. Unified structural descriptors and the fused Struct+Sent variant show the most stable behaviour across datasets.}
\label{fig:auroc_bars}
\end{figure*}

Figure~\ref{fig:auroc_bars} shows AUROC for structural and semantic variants across FEVER, SciFact, and WikiBio. Whereas Table~\ref{tab:main_results} reports absolute scores, the bar plot highlights relative gaps between descriptor families and illustrates the multi-scale nature of structural confidence.

\subsection{SciFact Heatmap}

\begin{figure*}[t]
\centering
\includegraphics[width=0.98\linewidth]{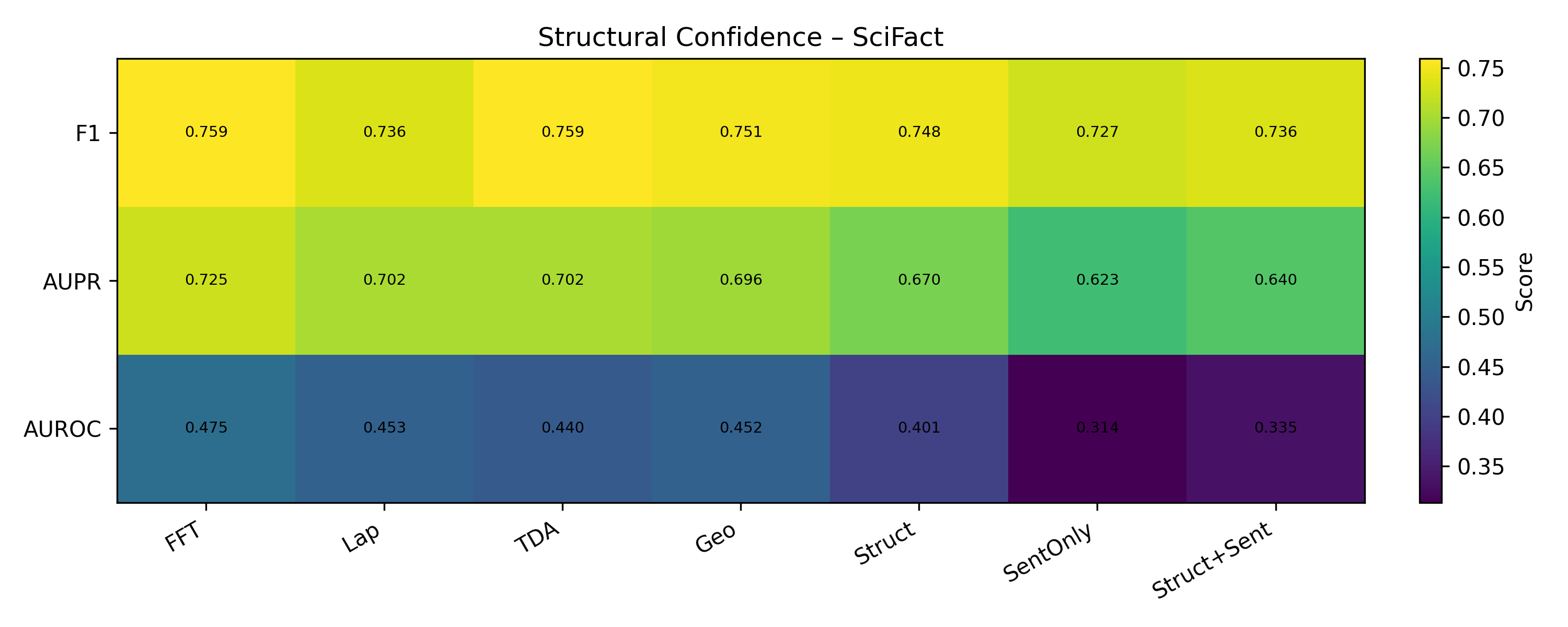}
\caption{Per-metric heatmap (AUROC and AUPR) for structural and semantic variants on SciFact. Semantic confidence exhibits a pronounced collapse relative to FEVER and WikiBio, while structural trajectories provide weaker but more stable signals across metrics.}
\label{fig:heatmap_scifact}
\end{figure*}

Figure~\ref{fig:heatmap_scifact} provides a per-metric heatmap (AUROC and AUPR) for the same variants on SciFact, illustrating the domain-shift effects discussed in Section~\ref{sec:crossdomain-results}. Semantic confidence performs strongly on FEVER and WikiBio but degrades sharply on SciFact, whereas structural variants retain weaker but more stable signals across metrics.

\section{Additional Ablations and Robustness Checks}
\label{app:ablation}

We report two additional ablations that complement the main text: PCA compression of the structural descriptor and a detailed energy analysis on FEVER.

\subsection{PCA Compression of the Structural Descriptor}
\label{app:ablation-pca}

To assess redundancy in the descriptor, we apply PCA to the full structural vector and retain a fixed number of principal components before training the classifier. Table~\ref{tab:ablation-pca} reports AUROC on FEVER and SciFact at several target dimensions.

Moderate compression (PCA-32) yields only small drops on FEVER, suggesting that a large fraction of the signal lies in a lower-dimensional subspace. On SciFact, where all methods struggle, aggressive compression degrades performance more noticeably. The full descriptor therefore provides a useful safety margin, especially in harder domains.

\begin{table}[t]
\centering
\begin{tabular}{lcccc}
\toprule
Dataset & Struct-only & PCA-32 & PCA-16 & PCA-8 \\
\midrule
FEVER   & $0.574$ & $0.552$ & $0.544$ & $0.534$ \\
SciFact & $0.460$ & $0.446$ & $0.412$ & $0.395$ \\
\bottomrule
\end{tabular}
\caption{Impact of PCA compression of the structural descriptor (two-scale setting, AUROC).}
\label{tab:ablation-pca}
\end{table}

\subsection{Detailed Energy Comparison on FEVER}
\label{app:ablation-energy}

\begin{table*}[t]
\centering
\begin{tabular}{lcccc}
\toprule
Method & \#Generations & Rel.\ FLOPs & Rel.\ Runtime & Rel.\ Memory \\
\midrule
SelfCheckGPT-style baseline & $5$ & $1.00$ & $1.00$ & $1.00$ \\
Structural Confidence (ours) & $1$ & $0.03$ & $0.04$ & $0.07$ \\
\bottomrule
\end{tabular}
\caption{Energy comparison on FEVER. Relative costs are normalised to a SelfCheckGPT-style baseline using $S=5$ stochastic generations per input. All methods rely on the same underlying LLM; differences arise solely from the number of generations and additional modules at inference time.}
\label{tab:ablation-energy}
\end{table*}

The main text reports average relative compute across datasets, normalised to Structural Confidence. Table~\ref{tab:ablation-energy} complements this with a single-dataset view on FEVER, normalised to a SelfCheckGPT-style baseline that performs multiple stochastic generations per input.

The results show the trade-off between multi-sample consistency and single-pass estimators. A SelfCheckGPT-style approach incurs substantial compute overhead, whereas Structural Confidence---using a single greedy generation and a lightweight gradient-boosted classifier—reduces FLOPs and runtime by over an order of magnitude. This supports the claim that structural signals can provide useful confidence estimates at a compute budget suitable for large-scale Web deployment.

\end{document}